# The Dem@Care Experiments and Datasets: a Technical Report

A. Karakostas[1], A. Briassouli[1], K. Avgerinakis[1], I. Kompatsiaris[1], M. Tsolaki[2]

[1] Information Technologies Institute, Centre for Research & Technology, Thessaloniki, Greece

{akarakos, abria, koafgeri, ikom}@iti.gr

[2] School of Medicine, Aristotle University of Thessaloniki, Thessaloniki, Greece

tsolakim1@gmail.com

### Abstract

The objective of Dem@Care is the development of a complete system providing personal health services to people with dementia, as well as medical professionals and caregivers, by using a multitude of sensors, for context-aware, multi-parametric monitoring of lifestyle, ambient environment, and health parameters. Multi-sensor data analysis, combined with intelligent decision making mechanisms, will allow an accurate representation of the person's current status and will provide the appropriate feedback, both to the person and the associated caregivers, enhancing the standard clinical workflow. Within the project framework, several data collection activities have taken place to assist technical development and evaluation tasks. In all these activities, particular attention has been paid to adhere to ethical guidelines and preserve the participants' privacy. This technical report describes shorty the (a) the main objectives of the project, (b) the main ethical principles and (c) the datasets that have been already created.

## The Dem@Care Project

The Dem@Care[1] project aspires to contribute to the timely diagnosis, assessment, maintenance and promotion of self-independence of people with dementia, by deepening the understanding of how the disease affects their everyday life and behaviour. It implements a multi-parametric closed-loop remote management solution that affords adaptive feedback to the person with dementia, while at the same time including clinicians into the remote follow-up, enabling them to maintain a comprehensive view of the health status and progress of the affected person. The system includes:

- a loop for people with dementia and their informal caregivers to monitor and assess their cognitive and behavioural status by integrating a multiplicity of wearable and in-situ sensors, enable time evolving context-sensitive profiling to support reactive and proactive care, and afford personalized and adaptive feedback.
- a loop for clinicians (neurologists, geriatric psychologists, geriatrists) specialized on dementia to provide objective observations regarding the health progression of the person with dementia and medication effectiveness, warn about trends closely related to dementia (e.g. apathy), and support preventive care decision-making and the adjustment of treatment recommendations.

---

[1] http://www.demcare.eu

**Data Collection**

The sensing (monitoring) technologies deployed in Dem@Care are tested on various benchmark datasets in order to evaluate the accuracy of the methodologies developed in the project. However, it is imperative to also test them in real world conditions, on elderly people, with or without dementia, for a fair and objective evaluation and in order to detect real life limitations and address them. For this reason, within the Dem@Care project, dementia-targeted data collection took place in the Greek Alzheimer's Association for Dementia and Related Disorders (GAADRD) in Thessaloniki, Greece. The GAADRD is a day center, offering advice, care and services to people with dementia and their families. The people recruited for the GAADRD dataset were elderly people, aged 65 and above, of both genders. The participant pool included people that were healthy and people suffering from conditions ranging from Mild Cognitive Impairment (MCI) to mild dementia, and in a few cases full-blown Alzheimer's Disease (AD). In most cases, people with mild cases of dementia were recruited, as they are still capable of carrying out activities of daily living (ADLs). The participants and their caregivers (relatives) were informed about the Dem@Care project and its goals and signed informed consent forms in order to participate in the data collection task. The multi-sensor data collected was then shared by project partners in a privacy-preserving manner and used in order to evaluate and validate sensing algorithms. In particular:

- audio data was used to extract correlations between vocal characteristics and cognitive state (dementia)
- static RGB (color) video data was used for recognition of ADLs and comparison with State of the Art (SoA) methods
- static RGB-D (color-depth) video data was used for online recognition of ADLs and comparison with State of the Art (SoA) methods
- wearable video data was used for object recognition, scene reconstruction, to aid in the more detailed recognition of ADLs.
- accelerometer data was used for physiological evaluation

Examples of the participants wearing the sensors in the realistic home-like recording environment set up in GAADRD can be seen in the figures below.

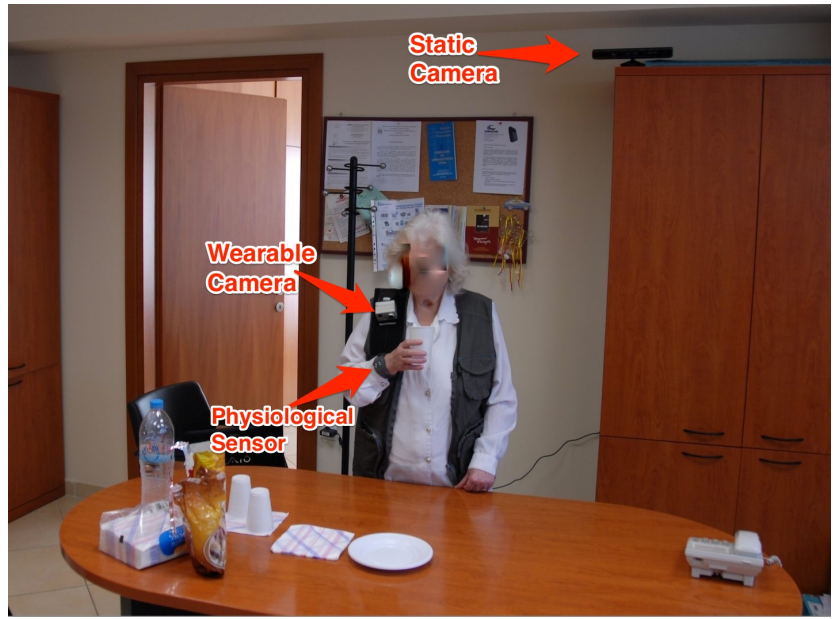

**Figure 1. Static camera, wearable camera and physiological sensor**

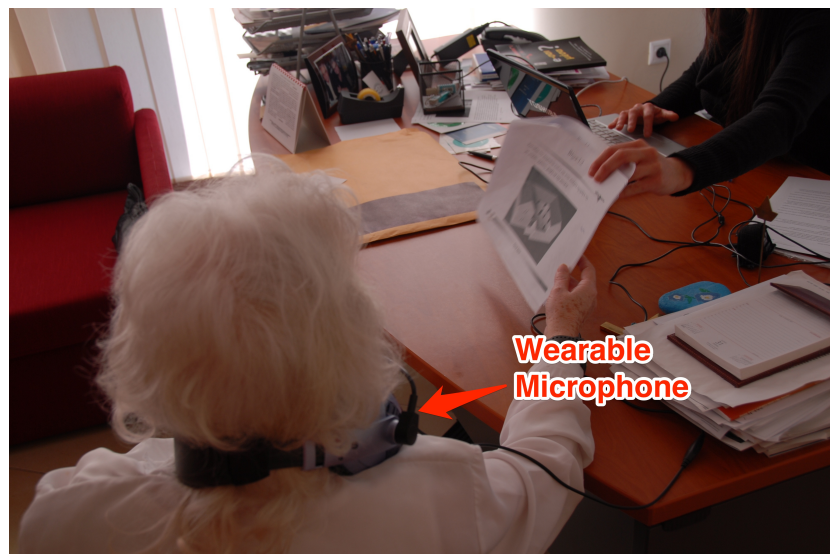

**Figure 2. Wearable microphone**

(a) Participant wearing the goPro camera on the vest, being recorded by the RGB-D camera placed on top of the bookcase. (b) Participant wearing a microphone during vocal tasks.

## Ethics Principles

Every participant who voluntarily participates in a Dem@Care experiment should sign[2] a consent form that mainly describes (a) the project objectives, (b) details about the experiment and the person's participation and (c) personal data protection issues.

Moreover, every research team that requests the Dem@Care datasets should follow specific instructions to complete an agreement form[3]. The main instructions are:

- Read the Dem@Care Datasets Terms of Use which oblige the research team, among other things, to:
    o not share this data with anyone else
    o not attempt to identify any of the individuals
- Choose which datasets they are going to use.
- Prepare a title and a short description (abstract) of their project.
- Send a mail to the Dem@Care coordinator (based on specific template) using their institution' s email with the agreement form completed.

## The Pilot Experiments and the Relevant Datasets

The pilot experiments of the Dem@Care project included 89 participants in various levels of severity of the cognitive / behavioral disturbances. The main datasets that emerged from these experiments are:

  o Video files from static, color-depth cameras (Asus RGB-D sensor, Kinect)
  o Video files from wearable camera (GoPro)
  o Audio files from microphone
  o Data from physiological sensors

## Acknowledgment

This work has been supported by the EU FP7 project Dem@Care: Dementia Ambient Care – Multi-Sensing Monitoring for Intelligent Remote Management and Decision Support under contract No. 288199.

---

[2] The consent form could also be signed by the patient's caregiver
[3] http://www.demcare.eu/results/datasets